# The fortresses of Ejin: an example of outlining a site from satellite images

**Amelia Carolina Sparavigna**
Department of Applied Science and Technology, Politecnico di Torino, Italy

*From 1960's to 1970's, the Chinese Army built some fortified artificial hills. Some of them are located in the Inner Mongolia, Western China. These large fortresses are surrounded by moats. For some of them it is still possible to see earthworks, trenches and ditches, the planning of which could have a symbolic meaning. We can argue this result form their digital outlining, obtained after an image processing of satellite images, based on edge detection.*

Ejin is a "banner" of the Inner Mongolia, that is, an administrative division of this Autonomous Region of China. The division got the name from the Ejin River, which is flowing near the ancient Khara-Khoto, the Black City, which is now deserted, capital of the Western Xia [1]. This region is a part of the large Juyan Lake Basin [2]. In the Figure 1 we can see the site of the Black City and another fortified place, partially destroyed by a flood.

The Ejin River flows through the Hexi Corridor, a valley which contained a portion of the Silk Road [3]. Outposts were built to recover and protect the traders moving on this road [4]. As discussed in Ref.5, it was during the first century BCE that the Chinese rulers started to create some fortified towns to protect the Corridor from the raids by Huns. Then a Hun ruler founded his own towns and later Chinese dynasties repeated the process. This sequence was possible by an abundance of water in this valley surrounded by arid regions. The towns were fed by cultivating oases and grassland was converted in cultivate land. However, this intensive use of the land for cultivation caused increasing erosion, which led to desertification and a gradual reduction of the river flow [5], and eventually, the towns along the Silk Road were abandoned. It seems that the loss of topsoil in this area has been recently intensified [6].

The satellite survey can help the archaeological studies of the Silk Road and the kingdoms along it [7,8], because the remote sensing can be used as a rapid procedure for detecting and localize some relevant surface and shallow-depth structures. Usually integrated in a geographic information system (GIS), the satellite survey has become an increasingly important tool for archaeologist [9]. If we consider the land around the Ejin river, being it a part of a corridor of the Silk Road, we can see that the outlines of ancient towns and fortresses are clearly visible in the satellite maps (see Fig.1). However, these are not the only interesting structures in this region.

Along the river, highly visible in the satellite maps because of their moats, there are other fortresses. These sites are not ancient, but they were built from 1960's to 1970's, by the Chinese Army [10,11]. Such fortresses consisted of a huge bunker in an artificial hill, or "mountain", 20 to 40 meters high and 250 to 400 meters wide. These artificial hills also assume the role of artillery observation and command posts. The forts included long-range artillery batteries, light anti-tank weapons, antiaircraft weapons, and other infantry weapons. The total strength of each of them was of about 300-500 people [11].

To localize them, it is possible to use Wikimapia, which is a GIS based on the Google Maps. The sites are labeled as mountains. In fact, since the artificial hills are surrounded by moats, they are quite visible in this desert land near the border between China and Mongolia.

Matching the appropriate scale to the level of details we want, we can see that the sites were connected by a railroad, in an overall system against enemies. Moreover each fortress is surrounded by satellite sites, which are increasing the size of the structure. Let us consider one on them, that at coordinates 41.766927N,100.73733E. It is shown in the Figure 2, which is obtained from a collage of Bing Maps. These maps are good for the processing that we want to do, to get the outline of all the structures, that is, the hill, satellite places and earthworks surrounding it (if we were interested to a finer scale, the Google Maps are more suitable). The upper image in the Figure 2 is processed in two manners gaining two output images: an image is obtained by applying the Sobel Edge Detector of GIMP [12], the other is obtained after applying AstroFracTool [13-15], which is a processing tool based on the calculus of the fractional gradient. These two images are then used as two layers to merge with GIMP in the lower image of Figure 2.

After processing, we see the outline of the military structure. There are five satellites sites, four of them having a triangular shape. Note that the hill surrounded by the moat is 600 m long and 500 m wide. The satellite sites are more than 100 meter wide. They are connected by undulated earthworks or trenches to the moat. All the military structure is covering the surface of a circle having a radius of 1 km. It is interesting to have the complete vision of this large area, because it is displaying a pentagonal planning. Of course, fortresses can have different shapes, depending on the available space in their location, but pentagonal shape is often chosen. In the ancient time, when the fortresses were walled structures, the pentagonal shape provided the best view of the adjacent sides with a low cost for materials and soldiers employed for the defense. For instance, the fortified citadel of Turin had a pentagonal shape [16].

Of course, in the case of the Ejin fortress, it was not a problem of the view from its walls. Supposing that the satellite sites hosted some artillery batteries, the shape could have be chosen to avoid a problem of crossfire. However, there is another possibility to consider, that the planning had a symbolic meaning too. Seen from the space, it resembled the shape of an ancient pentagonal fortress, especially when the ditches were full of water. Besides the pentagonal shape, some symbolic elements could have been included in the planning of the site. Let us look for example at Figure 3, where we can see a detail of one of the satellite triangles. The undulated line, connecting the triangle with the moat is crossing what looks like a flag.

In this paper we have shown that the processing of satellite imagery, besides enhancing the details of an image, can outline the structures observed in it. Therefore, the processed image can show the planning of the site and of its original structure. The same image can be suitable to be converted in a map suitable for historical or archaeological studies. As an example, we have discussed the fortresses of Ejin, in the Inner Mongolia. The outline of one of them revealed a complex planning that, probably, had a symbolic meaning too.

**References**
[1] http://en.wikipedia.org/wiki/Khara-Khoto
[2] http://en.wikipedia.org/wiki/Juyan_Lake_Basin
[3] http://en.wikipedia.org/wiki/Ruo_Shui, http://en.wikipedia.org/wiki/Ejin_River
[4] Yichun Xie, Robert Ward, Chuanglin Fang, Biao Qiao, The urban system in West China: A case study along the mid-section of the ancient Silk Road – He-Xi Corridor, Cities, Volume 24, Issue 1, February 2007, Pages 60–73
[5] Hou, Ren-zhi, Ancient city ruins in the deserts of the Inner Mongolia Autonomous Region of China, Journal of Historical Geography, 1985, 11 (3), pag.241-252
[6] Xiao-Gang Li, Feng-Min Li, Zed Rengel, Bhupinderpal-Singh, Zhe-Feng Wang, Cultivation effects on temporal changes of organic carbon and aggregate stability in desert soils of Hexi Corridor region in China, Soil and Tillage Research, Volume 91, Issues 1–2, December 2006, Pages 22–29

[7] A.C. Sparavigna, The road to the Loulan Kingdom, arXiv, 2012, arXiv:1210.5702, [physics.hist-ph]
[8] Lü HouYuan, Xia XunCheng, Liu JiaQi, Qin XiaoGuang, Wang FuBao, Yidilisi Abuduresule, Zhou LiPing, Mu GuiJin, Jiao YingXin, Li JingZhi, A preliminary study of chronology for a newly-discovered ancient city and five archaeological sites in Lop Nor, China, Chinese Sci Bull, 2010, Vol.55, pag.63–71.
[9] A.C. Sparavigna, Image Processing for the Enhancement of Satellite Imagery, in Image Processing: Methods, Applications and Challenges, edited by Vítor Hugo Carvalho, 2012, Nova Science Publishers, Inc. (USA), pp. 149-161. ISBN 9781620818442
[10] Information about the moated fortresses built by the Chinese Army are coming from Wikimapia (a GIS on Google Maps) and a Google Earth Community group moderated by Hill. http:// productforums.google.com / forum /#!msg / gec-military-moderated / TyhpEuCOIf0 / MlOtMTxQUR8J . At this address is reported that these are several place that can be seen in the satellite maps and the related KMZ files are given. It is also reported a paragraph from the book entitled "China Hands: Nine Decades of Adventure, Espionage, and Diplomacy in Asia" by James R. Lilley and Jeffrey Lilley, talking of them.
[11] Robin Grayson and Chimed-Eerdene Baatar, Remote sensing of cross-border routes between Mongolia and China, World Placer Journal, 2009, Volume 9, pag.48-118.
[12] GIMP, http://www.gimp.org/
[13] R. Marazzato and A.C. Sparavigna, Astronomical image processing based on fractional calculus: the AstroFracTool. arXiv, 2009, arXiv:0910.4637 [astro-ph.IM]
[14] A.C. Sparavigna, Fractional differentiation based image processing, arXiv, 2009, arXiv:0910.2381 [cs.CV]
[15] A.C. Sparavigna, Crater-like landform in Bayuda desert (a processing of satellite images), arXiv, 2010, arXiv:1008.0500 [physics.geo-ph]
[16] http://www.syler.com/SiegeWarfare/outside/digdowncitadel.html

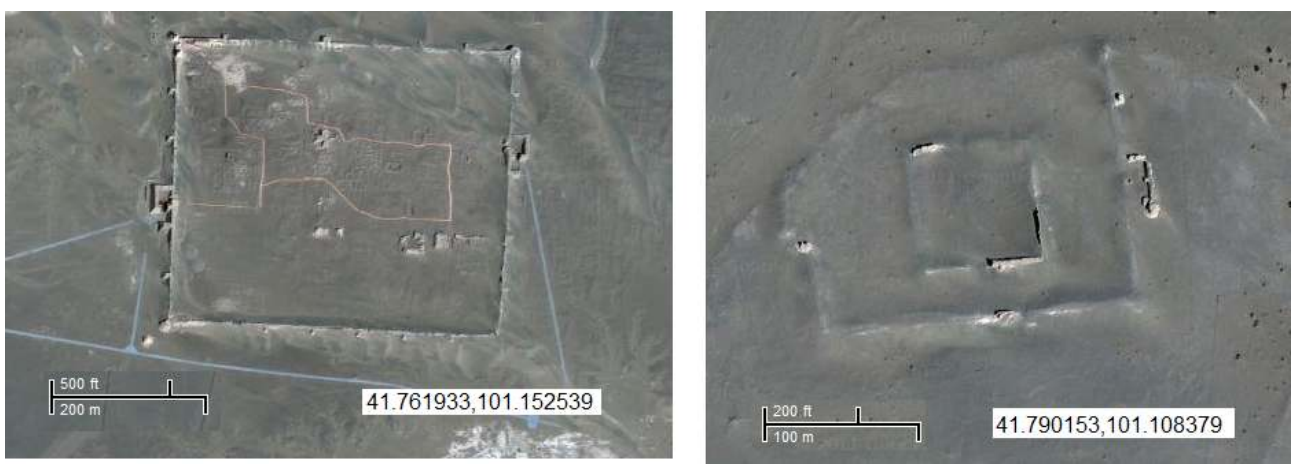

Fig.1 In these Google Maps we can see the site of Black City (on the left) and a fortified place, partially destroying by a flood (on the right).

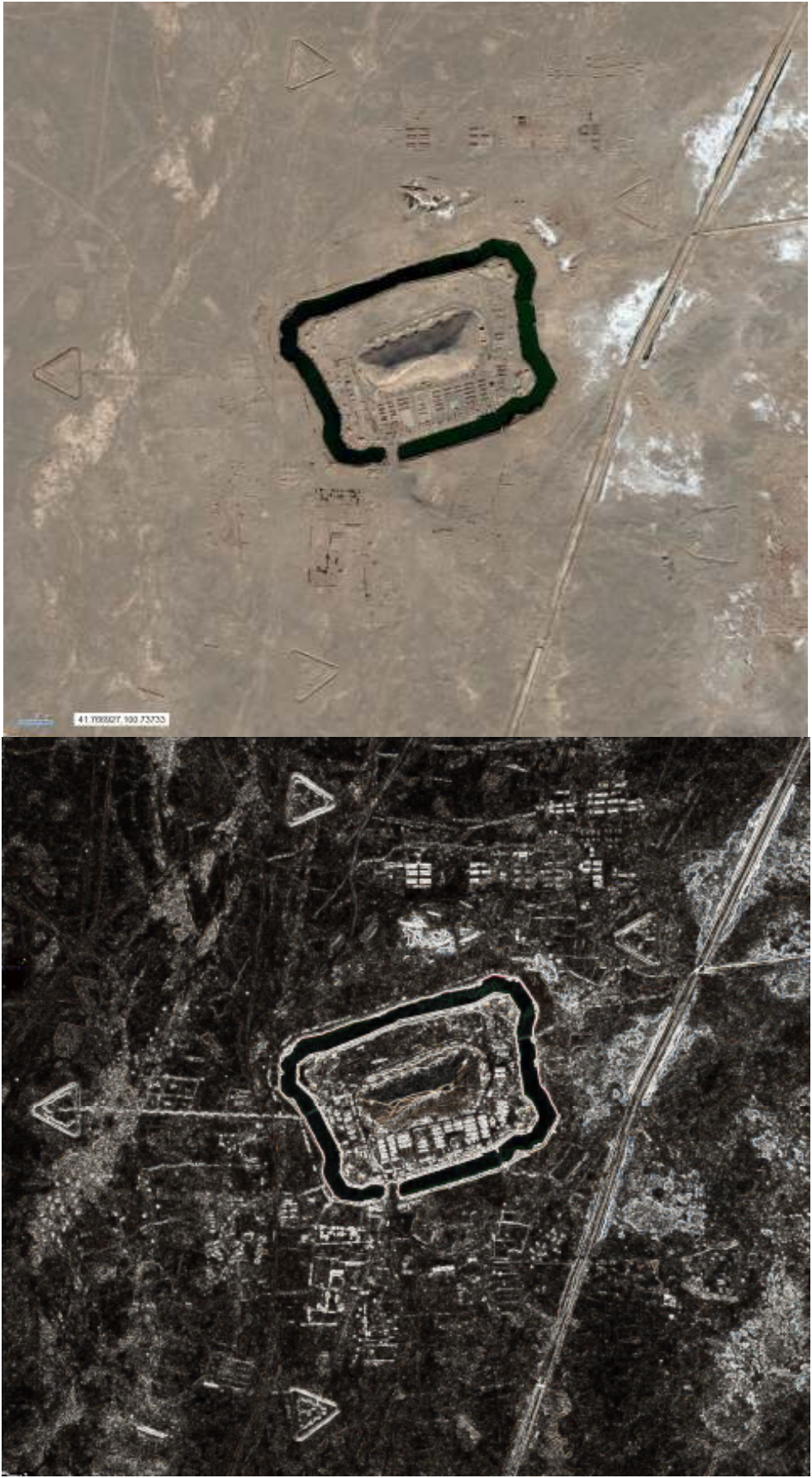

Fig.2 One of the fortresses in Ejin, an artificial hill surrounded by a moat. In the upper part, we see the site as shown in the maps. After processing it as discussed in the text, we obtain the outline of the complete site.

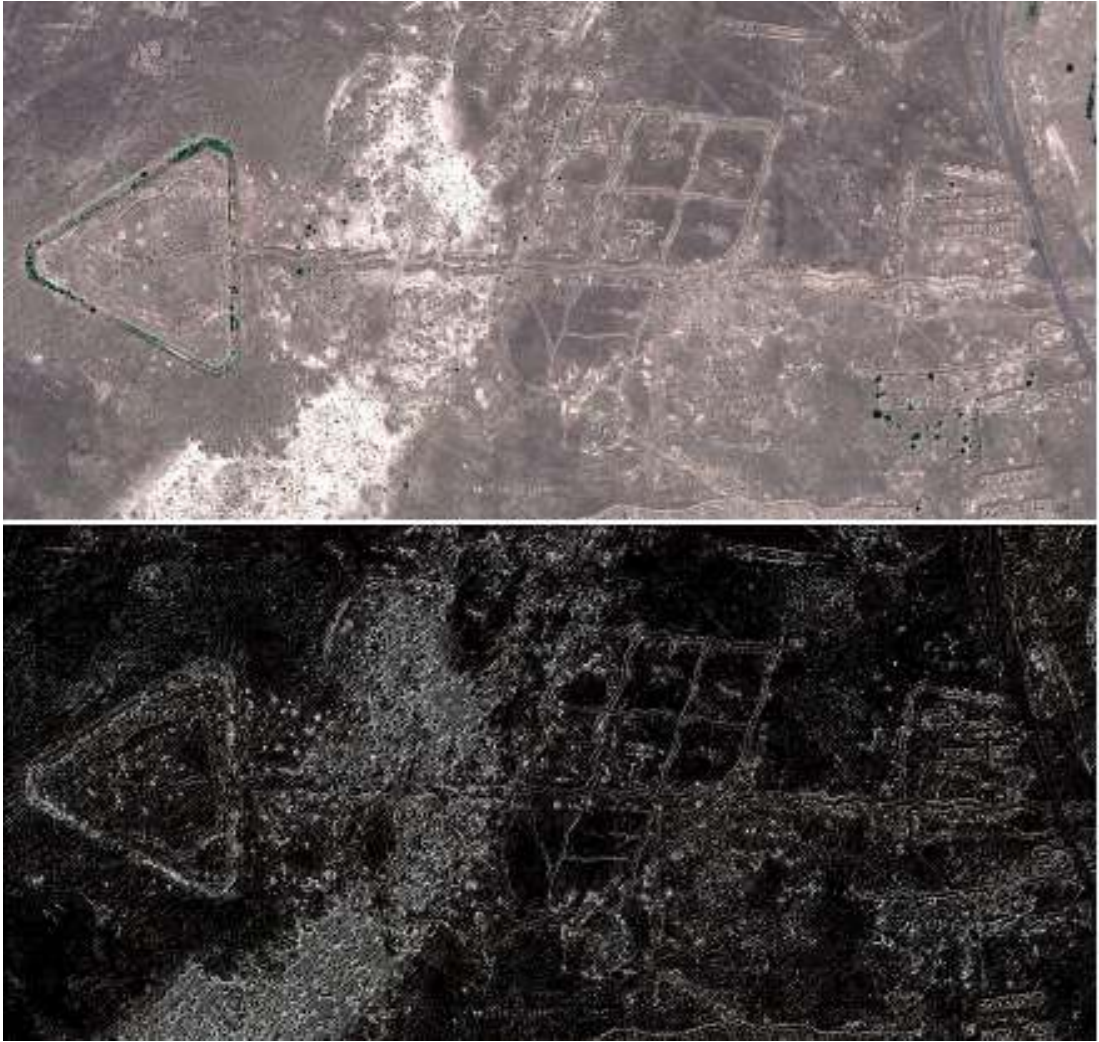

Fig.3 A detail of one of the fortresses in Ejin, as it is in the satellite images and after the processing.